\documentclass[11pt]{article}

\usepackage[final]{acl}

\usepackage{times}
\usepackage{latexsym}

\usepackage[T1]{fontenc}
\usepackage[utf8]{inputenc}
\usepackage{microtype}
\usepackage{inconsolata}

\usepackage{graphicx}
\usepackage{amsmath}
\usepackage{algorithm}
\usepackage{algpseudocode}
\usepackage{multirow}
\usepackage[dvipsnames, svgnames, x11names]{xcolor}
\usepackage{colortbl}
\usepackage{float}
\usepackage{hyperref}
\usepackage{amsthm}

\newcommand{\credit}{\mathrm{credit}}
\newcommand{\supp}{\mathrm{sup}}
\newcommand{\cov}{\mathrm{cov}}
\newcommand{\verifiedop}{\mathrm{verified}}
\newcommand{\stepof}{\mathrm{step}}
\newcommand{\Vtar}{\mathcal{V}_{\mathrm{tar}}}
\newcommand{\Utar}{\mathcal{U}_{\mathrm{tar}}}
\newcommand{\Etar}{\mathcal{E}_{\mathrm{tar}}}

\definecolor{linkblue}{HTML}{000099}
\hypersetup{
    colorlinks=true,
    linkcolor=linkblue,
    urlcolor=linkblue,
    citecolor=linkblue,
    pdfborder={0 0 0}
}
\usepackage{amsmath, amssymb, amsthm}
\usepackage{amsthm}

\usepackage{enumitem}
\usepackage{booktabs}
\usepackage{array}
\usepackage{tabularx}
\usepackage{subcaption}
\usepackage[most]{tcolorbox}
\newtcolorbox{promptbox}[1]{
  enhanced,
  colback=gray!5, colframe=gray!60!black,
  coltitle=white, fonttitle=\bfseries\small,
  title={#1},
  boxed title style={colback=gray!70!black},
  attach boxed title to top left={xshift=2mm, yshift=-2mm},
  left=4pt, right=4pt, top=4pt, bottom=4pt,
  fontupper=\small
}

\newcommand{\method}{STAMP}
\newcommand{\datasyn}{CHP}

\title{STAMP: Provenance-Guided Credit Assignment for Deep Search Agents}

\author{
  \textbf{Ke Xu\textsuperscript{1,2,*}},
  \textbf{Han Xu\textsuperscript{1,*,\dag}},
  \textbf{Xinran Chen\textsuperscript{1}},
  \textbf{Yuqian Wang\textsuperscript{1}},
  \textbf{Zhixuan Li\textsuperscript{1}},
\\
  \textbf{Xiaojian Liu\textsuperscript{1}},
  \textbf{Changwo Wu\textsuperscript{1}},
  \textbf{Jianqiang Xia\textsuperscript{1}},
  \textbf{Yuchen Li\textsuperscript{1}}
\\
\\
  \textsuperscript{1}Baidu Inc.,
  \textsuperscript{2}Peking University
\\
  \small{
    \texttt{xuke59@stu.pku.edu.cn},
    \texttt{xhbj66@gmail.com}
  }
}

\begin{document}
\maketitle
\begin{abstract}
Reinforcement learning for deep-search agents has largely focused on trajectory-level scoring---outcome correctness, citation-aware rewards, and evidence coverage. Yet the actions that expose supporting documents receive no targeted credit, a gap we call the \emph{reward-credit mismatch}.
We propose \method{}, in which a reference-based verifier judges whether each cited document supports an entity or relation in a training-time evidence graph, and first-exposure attribution traces each supported citation back to the action that first surfaced it.
This step credit is injected through sign-preserving advantage modulation, which redistributes advantage across steps without changing the trajectory-level reward or the relative ranking of trajectories within each group.
On BrowseComp, BrowseComp-ZH, and xbench-DS, \method{} improves the GRPO baseline by \textbf{+2.0}/\textbf{+5.5}/\textbf{+3.0} points under matched SFT initialization, training data, and search tools, and composes with both outcome-only and citation-rubric base rewards.
Component ablations confirm that the provenance-based credit signal and the sign-preserving advantage modulation each contribute to the gains.

\end{abstract}

\section{Introduction}
Deep search agents discover hidden entities, verify relations across webpages, and assemble cited evidence to support a final answer~\cite{yao2023reactsynergizingreasoningacting,openai2025deepresearchsystem}.
Reinforcement learning has become the dominant training paradigm for improving such agents, but long-horizon search exposes a distinction that trajectory-level training tends to blur: trajectory-level scoring of a rollout differs from step-level credit assignment to the actions that produced it~\cite{zhang2026landscapeagenticreinforcementlearning}.

\begin{figure}[t]
  \centering
  \begin{subfigure}[t]{0.44\textwidth}
    \centering
    \includegraphics[width=\linewidth]{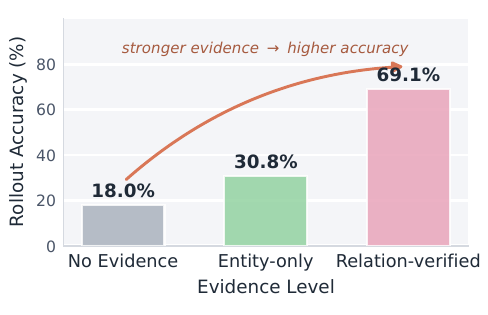}
    \caption{Accuracy by evidence level.}
    \label{fig:motivation_funnel}
  \end{subfigure}
  \hfill
  \begin{subfigure}[t]{0.44\textwidth}
    \centering
    \includegraphics[width=\linewidth]{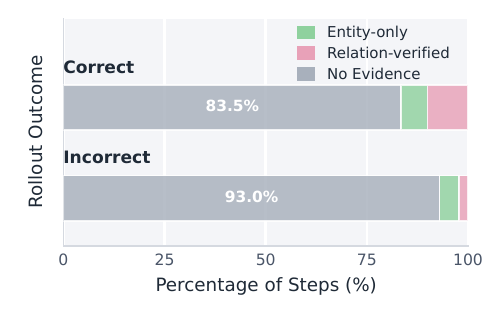}
    \caption{Per-step evidence credit.}
    \label{fig:motivation_credit}
  \end{subfigure}
\caption{Evidence in deep-search rollouts. 
(a) Accuracy by the highest evidence level reached: none, entity-only, or relation-verified. 
(b) Per-step evidence credit by trajectory correctness, showing that evidence-producing steps are sparse and occur in both correct and incorrect rollouts.}
  \label{fig:motivation_all}
 \vspace{-0.4cm}
\end{figure}

Most recent progress targets the scoring side: outcome correctness, or how to score final answers, citations, and evidence coverage at the trajectory level~\cite{jin2025searchr1trainingllmsreason,gao2025turnsunlockinglonghorizonagentic,Li2025WebSailorNS}.
Richer trajectory-level rewards---including rubric-based scoring, citation-aware rewards, entity- or relation-level evidence rewards, and multi-turn outcome supervision~\cite{zhang2026chainingevidencerobustreinforcement,zhao2026repurposingsyntheticdatafinegrained}---have substantially improved deep-search agents, especially when paired with synthetic data and cold-start trajectories.
Our analysis is consistent with this direction, but suggests that the value of a rollout is largely determined by the evidence its steps actually produce: Fig.~\ref{fig:motivation_funnel} reports accuracy rising sharply from no target evidence (18.0\%) to entity-only evidence (30.8\%) to relation-verified evidence (69.1\%). Outcome-level rewards score whether such evidence ends up cited, but do not directly reward the actions that exposed it.

The complementary side---step-level evidence, namely which search or read actions first expose target entities and verify relations---remains far less exploited.
In standard outcome-supervised RL, the trajectory's correctness verdict is the only signal that survives group-relative normalization, and the resulting advantage is broadcast uniformly to every action token~\cite{shao2025deepseekmathv2selfverifiablemathematicalreasoning,guo2025segmentpolicyoptimizationeffective}.
This broadcast hides a strong mismatch between outcomes and evidence-producing actions: Fig.~\ref{fig:motivation_credit} shows that 83.5\% of steps in correct rollouts produce no evidence at all, while 7.0\% of steps in incorrect rollouts still expose useful entity- or relation-level evidence.
We call this form of credit-assignment gap the \textbf{reward-credit mismatch}: the outcome channel captures whether a trajectory succeeds, but the evidence channel---which actions expose the supporting documents---remains an underused supervision signal.

One remedy is to inject step-level reward bonuses, but in standard outcome-supervised setups---including GRPO---trajectories receive a single scalar reward; per-step bonuses are summed into that scalar and broadcast uniformly, eliminating the temporal localization they were intended to provide.
Learned process reward models could deliver local feedback in principle~\cite{yin-etal-2025-dynamic,choudhury2025processrewardmodelsllm}, but applying them to open-web search requires judging arbitrary intermediate actions in a large, evolving action space, often without an explicit reference for what evidence the action should expose.

Instead, we ground credit in citation provenance recorded by the trajectory.
Verification reduces to a reference-based judgment---whether a cited document supports a given entity or constraint from the training-time evidence graph---thus avoiding scoring the action itself.
Existing citation-aware methods consume citations as input to a trajectory-level reward~\cite{nakano2022webgptbrowserassistedquestionansweringhuman,zhang2026chainingevidencerobustreinforcement}; we instead use them as temporal provenance pointers, tracing each supporting document back to the action that first exposed it.
This is possible in our setting because agents emit citations and our verifiable data-synthesis pipeline already constructs a training-time evidence graph for each query.

We propose \textbf{\textsc{STAMP}} (\textbf{S}tep-level \textbf{T}race-guided \textbf{A}dvantage \textbf{M}odulation with \textbf{P}rovenance), which addresses the reward--credit mismatch by decoupling two questions: what evidence a trajectory verifies, and which action first made that evidence available.
The resulting step credit is injected through sign-preserving advantage modulation, a form of step-level advantage redistribution that leaves the trajectory-level reward and the induced trajectory ranking untouched. The evidence graph is used only at training time; at inference, the agent must rediscover entities, relations, and supporting documents through live search.

On Qwen3-30B-A3B-Thinking-2507~\cite{qwen3technicalreport}, we evaluate \textsc{STAMP} in a live-web deep-search setting on BrowseComp~\cite{wei2025browsecompsimplechallengingbenchmark}, BrowseComp-ZH~\cite{zhou2025browsecompzhbenchmarkingwebbrowsing}, and xbench-DS~\cite{xbench2025xbench}. All controlled baselines share identical SFT initialization, training data, and search tools. On top of outcome-only GRPO, \textsc{STAMP} improves accuracy by \textbf{+2.0}/\textbf{+5.5}/\textbf{+3.0} points on the three benchmarks.
The gains are largest for outcome-only and citation-rubric rewards and smaller for entity-only reward, suggesting that citation-provenance credit is most useful when the trajectory-level reward does not already densely supervise entity recovery. Component ablations confirm that verification quality, first-exposure attribution, bounded modulation, and negative-advantage attenuation each contribute to the final performance.

Our contributions are as follows: We empirically characterize the reward-credit mismatch in deep-search RL: 83.5\% of steps in correct rollouts produce no target evidence, while 7.0\% of steps in incorrect rollouts still expose useful entity- or relation-level evidence. We propose \textsc{STAMP}, which turns citation provenance into verifiable step credit via a reference-based verifier and first-exposure attribution, then injects this credit through sign-preserving advantage modulation while leaving the trajectory-level reward and ranking untouched. \textsc{STAMP} improves BrowseComp, BrowseComp-ZH, and xbench-DS by \textbf{+2.0}/\textbf{+5.5}/\textbf{+3.0} points over the GRPO baseline under matched SFT initialization, training data, and tools, and composes with outcome-only and citation-rubric rewards.

\section{Methodology}
\label{sec:method}

\begin{figure*}[t]
  \centering
  \includegraphics[width=\textwidth]{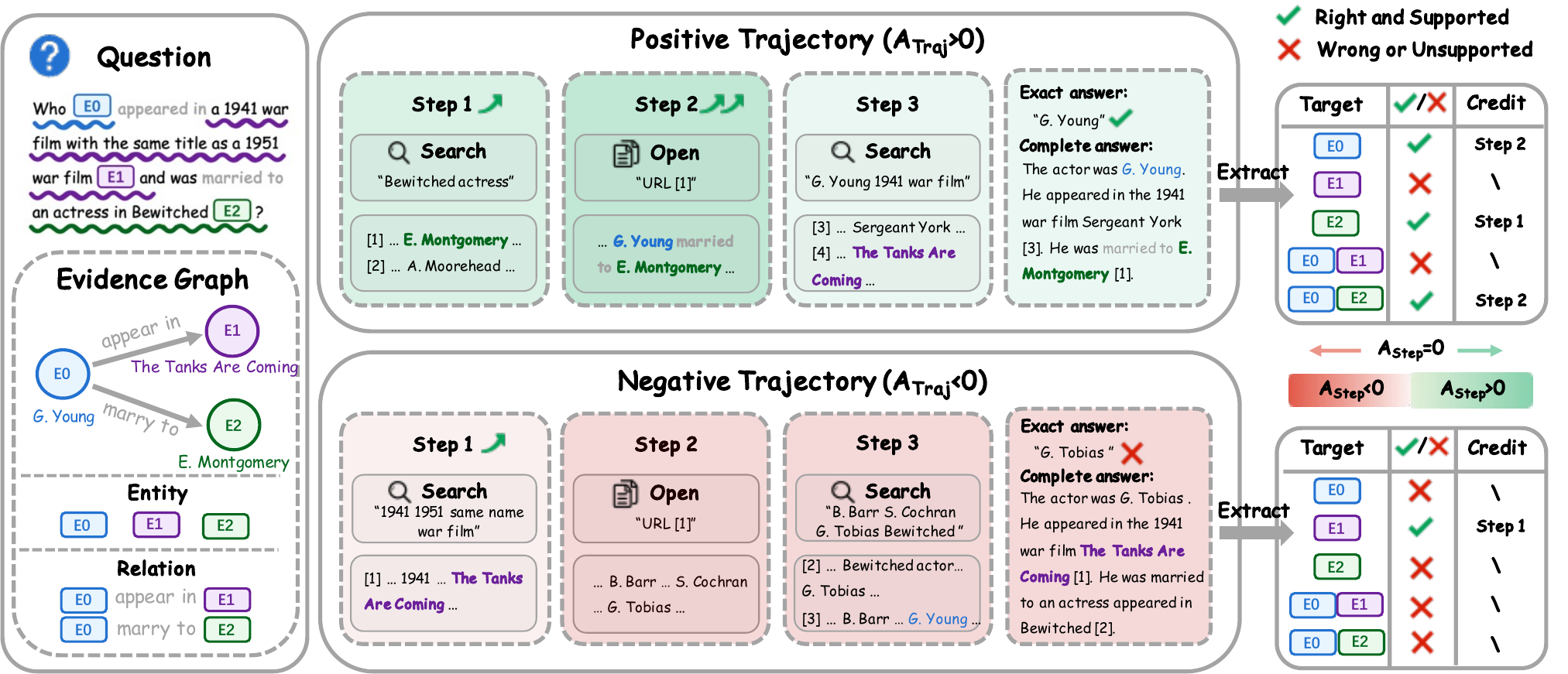}
  \caption{Overview of \textsc{STAMP}. Citation provenance traces verified evidence back to its earliest exposure step, converting document-level support into step-level credit. Sign-preserving advantage modulation then amplifies credited steps in positive trajectories ($A_{\mathrm{traj}}>0$) and shields them from uniform penalization in negative ones ($A_{\mathrm{traj}}<0$), while uncredited steps retain the original trajectory advantage.}
  \label{fig:framework}
  \vspace{-0.4cm}
\end{figure*}

\textsc{STAMP} (\textbf{S}tep-level \textbf{T}race-guided \textbf{A}dvantage \textbf{M}odulation with \textbf{P}rovenance) addresses the reward--credit mismatch in outcome-only training by decoupling two questions: \emph{what} evidence a trajectory verifies, and \emph{which} action first made it available.
Given a reference-anchored verifier that judges cited documents against the training-time evidence graph (\S\ref{sec:reward}), \textsc{STAMP} (i) uses \textbf{citation provenance} to map each supporting document to its earliest exposure step, yielding bounded step-level credit (\S\ref{sec:credit}), and (ii) injects this credit through \textbf{sign-preserving advantage modulation} that leaves the trajectory-level reward $R(\cdot)$ untouched (\S\ref{sec:optimization}).
The design is orthogonal to the choice of trajectory-level reward.

\subsection{Preliminaries: Deep Search with Evidence Traces}
\label{sec:preliminaries}

\paragraph{Task Formulation}
We formulate deep search as an episodic decision process.
At step $t$, the agent observes history $h_t=(q,a_1,o_1,\ldots,a_{t-1},o_{t-1})$ and issues an action $a_t$ such as a search query, page-reading request, or final cited response; the environment returns observation $o_t$.
An episode ends with a final answer $r$ and cited URLs, yielding trajectory $\tau=(a_1,o_1,\ldots,a_T,r)$.
Outcome-supervised training assigns a terminal reward $R(\tau)$ from final-answer correctness, leaving intermediate search actions with undifferentiated credit.

\paragraph{Training-Time Evidence Graph}
Each training query $q$ is paired with a training-time evidence graph $\mathcal{G}=(\mathcal{V},\mathcal{E})$.
Nodes $\mathcal{V}$ are task-relevant entities, with target entities $\Vtar\subseteq\mathcal{V}$ denoting \emph{hidden} entities that are not revealed in $q$ and must be recovered through search; entities mentioned in $q$ are excluded from $\Vtar$ by construction and never enter the entity-credit budget.
Edges $(e_i,e_j)\in\mathcal{E}$ are single-hop factual relations instantiated as atomic, single-document-verifiable natural-language constraints $c_{ij}$; $\Etar\subseteq\mathcal{E}$ denotes the target constraints used for verification.
The graph is only a training-time scaffold for reward computation and credit assignment, not an exhaustive set of valid evidence paths; at inference time, the agent receives only the query.
Training data synthesis details are provided in Appendix~\ref{sec:chp_details}.

\paragraph{Standard GRPO}
GRPO~\citep{shao2024deepseekmathpushinglimitsmathematical} samples a group of $G$ rollouts $\{\tau_1, \ldots, \tau_G\}$ from $\pi_\theta$ given $q$, scores each with $R_i = R(\tau_i)$, and produces a trajectory-level advantage
\begin{equation}
  A_i = \frac{R_i - \mu}{\sigma + \epsilon},
\label{eq:grpo_advantage}
\end{equation}
where $\mu, \sigma$ are the group mean and standard deviation of $\{R_i\}$ and $\epsilon$ guards zero-variance groups; $A_i$ is then broadcast uniformly to every action token of $\tau_i$.

\subsection{Evidence Verification}
\label{sec:reward}

For each target edge $(e_i,e_j)\in\Etar$ with constraint $c_{ij}$, define a verification unit $u_k=(e_i,e_j,c_{ij})$ and let $\Utar$ be the set of such units.
Let $\mathcal{J}$ be a binary verifier over cited document contents that judges support for a \emph{specific} entity or constraint---not general topical relevance---against a reference drawn from the evidence graph.
It is invoked post-hoc rather than inside the policy update.
We instantiate $\mathcal{J}$ as an LLM-based extractor-matcher; verifier prompts and validation accuracy are reported in Appendix~\ref{sec:prompts}.
After URL canonicalization, $\mathcal{A}_t$ denotes the documents whose URL first becomes available at step $t$ (via a query result or read action), and $\mathcal{C}(\tau)$ denotes the documents cited in $r$ whose URL matches one observed in $\tau$; unresolved citations are discarded.

For query-mentioned entities (excluded from $\Vtar$ by construction), we set $\cov(e)=1$ so units anchored on a revealed endpoint remain groundable; such units generate no entity credit, and the relation gate $\verifiedop(u_k)=1$ (Eq.~\ref{eq:evidence_indicators}) still requires the hidden endpoint and the constraint to be supported by cited documents.
For target entities and grounded verification units, the verifier induces supporting document sets
\begin{equation}
\begin{aligned}
\mathcal{D}(e) &= \{d\in\mathcal{C}(\tau): \supp(e,d)=1\}, \\
\mathcal{D}(u_k) &= \{d\in\mathcal{C}(\tau): \supp(c_{ij},d)=1\}.
\end{aligned}
\label{eq:evidence_support_sets}
\end{equation}
The corresponding indicators are
\begin{equation}
\begin{aligned}
\cov(e) &= \mathbf{1}[\mathcal{D}(e)\neq\emptyset], \\
\cov(c_{ij}) &= \mathbf{1}[\mathcal{D}(u_k)\neq\emptyset], \\
\verifiedop(u_k) &= \cov(e_i)\,\cov(e_j)\,\cov(c_{ij}).
\end{aligned}
\label{eq:evidence_indicators}
\end{equation}

\subsection{Credit from Citation Provenance}
\label{sec:credit}

Having defined \emph{what} evidence the trajectory verified (\S\ref{sec:reward}), we now ask \emph{which} action receives credit for it.
We attribute credit only to actions that first expose evidence the trajectory later verifies, leaving all other intermediate actions unjudged.

\paragraph{Evidence-to-Document Credit}
Each grounded hidden target entity ($e\in\Vtar$ and $\cov(e)=1$) generates entity credit, and each verified unit ($\verifiedop(u_k)=1$) generates relation credit, even when one endpoint entity is already revealed in the query.
Entity and relation units use the same per-unit weight $\delta$; the difference between the two channels lies in their trigger conditions rather than in per-unit weighting.
By Eqs.~\ref{eq:evidence_support_sets}--\ref{eq:evidence_indicators}, $\mathcal{D}(e)$ and $\mathcal{D}(u_k)$ are nonempty whenever the corresponding indicator equals $1$, so the divisions below are well-defined.
Document-level credit is distributed uniformly across each supporting document set:
\begin{align}
  \credit_{\mathrm{ent}}(d,e)
  &= \frac{\delta}{|\mathcal{D}(e)|}, \quad d\in\mathcal{D}(e),
  \label{eq:credit_ent} \\
  \credit_{\mathrm{rel}}(d,u_k)
  &= \frac{\delta}{|\mathcal{D}(u_k)|}, \quad d\in\mathcal{D}(u_k),
  \label{eq:credit_rel}
\end{align}
when $e\in\Vtar$ with $\cov(e)=1$ and when $\verifiedop(u_k)=1$, respectively, and $0$ otherwise.
Uniform within-set distribution avoids introducing a learned weighting over documents, keeping attribution reproducible given verifier outputs.
A document supporting both an entity and a verified relation receives credit from both channels, since the two channels measure complementary contributions (identification vs.\ relation establishment).
Credit attribution applies to all trajectories regardless of outcome; low-reward trajectories may still contain steps that retrieved genuinely useful evidence, and \S\ref{sec:optimization} ensures such steps are not uniformly penalized when the group-relative advantage is negative.

\paragraph{Provenance Map to Search Steps}
To make temporal localization reproducible from the trajectory log, we attribute each supporting document $d$ to the earliest action that makes it available to the agent.
Recalling the document set $\mathcal{A}_t$ defined in \S\ref{sec:reward}, we define the exposure step of document $d$ as:
\begin{equation}
  \stepof(d) = \min \big\{t : d \in \mathcal{A}_t\big\}.
  \label{eq:step_attr}
\end{equation}
All credit assigned to $d$ is transferred to this exposure step; later actions that re-encounter the same document receive no extra credit.
Earliest-exposure is a deliberate inductive bias rather than a default choice: it credits the action that first \emph{discovers} a supporting document, whereas a read-time alternative $\stepof(d)=\min\{t:a_t\text{ reads }d\}$ would credit the action that \emph{consumes} it.
In deep search, query formulation is the dominant bottleneck---many queries fail to surface any supporting document at all---so we route credit to discovery actions; learned alternatives (LLM-judged attribution, information-gain estimation) trade off this log-reproducibility for a separate attribution model and are left to future work.

\paragraph{Bounded Step Credit}
The total credit for step $t$ accumulates entity and relation contributions from all documents first exposed at that step:
\begin{equation}
\begin{aligned}
  E_t &= \sum_{\substack{e \in \Vtar\\ \cov(e)=1}}
         \sum_{\substack{d \in \mathcal{D}(e)\\ \stepof(d)=t}}
         \credit_{\mathrm{ent}}(d,e), \\
  U_t &= \sum_{\substack{u_k \in \Utar\\ \verifiedop(u_k)=1}}
         \sum_{\substack{d \in \mathcal{D}(u_k)\\ \stepof(d)=t}}
         \credit_{\mathrm{rel}}(d,u_k), \\
  \credit_t &= \min\!\big(C,\; E_t + U_t\big).
\end{aligned}
\label{eq:step_credit}
\end{equation}
The cap $C \in [0, 1)$ prevents any single step from dominating the gradient signal; together with $\credit_t \in [0, C]$ it ensures sign preservation in the modulation operator of \S\ref{sec:optimization}.
The cap is in practice active only when many evidence units concentrate on a single step, which can happen when a broad query coincidentally surfaces many relevant URLs at once.
Define the (uncapped) trajectory evidence budget as $B(\tau) := \sum_t (E_t + U_t)$.
Each grounded hidden target entity ($\cov(e)=1$) contributes $\delta$ in total across the trajectory, and each verified unit ($\verifiedop(u_k)=1$) contributes $\delta$; ungrounded entities and unverified units contribute $0$.
Combined with the monotone, dominated capped accumulator $\credit_t = \min(C, E_t + U_t)$, and writing $n_e(\tau) := |\{e\in\Vtar:\cov(e)=1\}|$ and $n_u(\tau) := |\{u_k\in\Utar:\verifiedop(u_k)=1\}|$ for the grounded-entity and verified-unit counts, we obtain
\begin{equation}
\begin{aligned}
  \sum_t \credit_t \;&\le\; B(\tau) \;=\; \big(n_e(\tau)+n_u(\tau)\big)\,\delta \\
  &\le\; (|\Vtar|+|\Utar|)\,\delta.
\end{aligned}
\label{eq:credit_budget}
\end{equation}
The bound depends only on the size of the target evidence graph, not on the trajectory length $T$; trajectories that take more steps distribute the same credit budget more sparsely rather than accumulating more total modulation.

\subsection{Sign-Preserving Advantage Modulation}
\label{sec:optimization}

Once provenance credit is localized to action steps, we use it to redistribute optimization pressure within the trajectory-level signal produced by GRPO, without introducing a separate local reward.
We let $m$ index generated action tokens and write $t(m)$ for the action step containing token $m$, and abbreviate the per-step credit of trajectory $\tau_i$ as $\credit_t^{(i)}$ (the quantity defined in Eq.~\ref{eq:step_credit}).

\paragraph{Sign-Preserving Modulation}
\textsc{STAMP} keeps $R(\cdot)$ and the group statistic in Eq.~\ref{eq:grpo_advantage} untouched and modulates the broadcast at the action-step level.
A trajectory-summed credit bonus $R'_i = R_i + \sum_{t'} \credit_{t'}^{(i)}$ would be rebroadcast uniformly by GRPO's group normalization and erase temporal localization, whereas modulation acts on $A_i$ at the step level, preserving per-step granularity and the trajectory ranking induced by $R(\cdot)$.
Concretely, we define the \emph{sign-preserving modulation operator}
\begin{equation}
  \mathcal{M}_c(A) := A\,\bigl(1 + \mathrm{sign}(A)\,c\bigr),
\label{eq:modulation_op}
\end{equation}
with $c\in[0,C]$ and cap $C\in[0,1)$, and apply it at the step level using the citation-provenance credit $\credit_t^{(i)}$, which depends only on citation provenance and verifier outputs, not on $A_i$ or $R_i$.
This gives $\widetilde{A}_{i,t}=\mathcal{M}_{\credit_t^{(i)}}(A_i)$, and all action tokens $m$ in the same step share the modifier $A^*_{i,m}=\widetilde{A}_{i,t(m)}$.
We adopt the multiplicative form over two alternatives sharing the same amplify-positive / attenuate-negative semantics:
\begin{itemize}
\setlength{\itemsep}{2pt}
\setlength{\topsep}{2pt}
\item \emph{Multiplicative} $A(1+\mathrm{sign}(A)c)$: the multiplier $1\pm c\in[1{-}C,\,1{+}C]$ is bounded away from zero for any $C<1$, so sign is preserved unconditionally; modulation scales with $|A|$ and vanishes as the group becomes homogeneous.
\item \emph{Additive} $A+\mathrm{sign}(A)\,c\,A_{\mathrm{ref}}$: sign preservation requires the trajectory-dependent condition $|A|>c\,A_{\mathrm{ref}}$ and may fail on near-consensus rollouts; this is equivalent to step-level reward shaping injected at the advantage layer.
\item \emph{Power} $\mathrm{sign}(A)|A|^{1-c}$: amplifies when $|A|<1$ but attenuates when $|A|>1$, contradicting the intended semantics in roughly half of the cases.
\end{itemize}
Among these three forms, the multiplicative form uniquely combines unconditional sign preservation with $|A|$-scaled modulation: the operator satisfies $\mathrm{sign}(\mathcal{M}_c(A))=\mathrm{sign}(A)$ and $|\mathcal{M}_c(A)|\in[(1{-}C)|A|,(1{+}C)|A|]$ for all $c\in[0,C]$, so credit only rescales optimization pressure and cannot reverse the direction set by the trajectory-level signal.
For positive-advantage trajectories, evidence-producing steps receive amplified reinforcement; for negative-advantage trajectories, those steps are attenuated by $1-c\in[1{-}C,\,1]$ rather than converted into positive supervision, so a failed trajectory remains penalized but its evidence-producing search actions are not penalized as if they were uninformative.
When $A_i=0$ the modulation vanishes by construction, so groups whose rollouts share the same reward provide no advantage-layer signal.

\paragraph{Gradient and Objective}
With $g_{i,m}(\theta)=\nabla_\theta\log\pi_\theta(a_{i,m}\mid h_{i,m})$, the modulated update reads
\begin{align}
A^*_{i,m}g_{i,m}(\theta)
&= \lambda_{i,m}\, A_i\, g_{i,m}(\theta), \\
\lambda_{i,m}
&= 1+\mathrm{sign}(A_i)\,\credit_{t(m)}^{(i)},
\label{eq:grad_redistribution}
\end{align}
with $\lambda_{i,m}\in[1{-}C,\,1{+}C]$, so every token gradient is rescaled but never sign-flipped.
Thus, \textsc{STAMP} reweights per-token learning pressure within a trajectory rather than redefining the learning target.
Combined with Eq.~\ref{eq:credit_budget}, the trajectory-level deviation it introduces is bounded by the evidence-graph size rather than by trajectory length.
The resulting clipped GRPO objective replaces the trajectory-level advantage broadcast to action tokens with $A^*_{i,m}$:
\begin{equation}
  \mathcal{L}(\theta) = -\mathbb{E}\!\left[\sum_{i,m} \mathbb{I}_{i,m}
  \min\!\bigl(\rho_{i,m} A^*_{i,m},\; \bar{\rho}_{i,m} A^*_{i,m}\bigr)\right],
\label{eq:stamp_loss}
\end{equation}
where $\bar{\rho}_{i,m}=\mathrm{clip}(\rho_{i,m}, 1{-}\varepsilon, 1{+}\varepsilon)$, $\rho_{i,m}$ is the generated-token importance ratio, and $\mathbb{I}_{i,m}$ masks non-action tokens.
The PPO/GRPO clipping is preserved unchanged; \textsc{STAMP} modulates only the advantage signal.
Training proceeds in four post-rollout steps: verify cited documents against the evidence graph, attribute supporting documents to exposure steps, compute step credits, and apply sign-preserving modulation during the GRPO update.
All verification and attribution are performed after rollout completion, so \textsc{STAMP} adds no cost during rollout generation; the overhead is limited to parallel post-hoc verification.

\section{Experiments}
\label{sec:exp}

\subsection{Setup}
\label{sec:setup}
\paragraph{Tools and Environment}
The agent interacts with the live web using two tools. The \texttt{search} tool queries the Serper API~\cite{serper} and returns ranked snippets with their corresponding URLs, while the \texttt{open} tool uses the Jina API~\cite{jina} to extract page content from a specified URL.

\paragraph{Training Setup}
The base model is Qwen3-30B-A3B-Thinking-2507~\cite{qwen3technicalreport}.
We synthesize bilingual training data from Wikipedia (Appendix~\ref{sec:chp_details}): 3000 queries for SFT (2000 EN, 1000 ZH) and 2,600 for RL (2,100 EN, 500 ZH).
Our training process includes cold-start SFT and subsequent RL.
For cold-start SFT, we leverage Seed~Pro~2.0~\cite{seed2026seed20} to generate 1,500 high-quality search trajectories through reject sampling on the SFT query set, then fine-tune the base model for 1 epoch with a batch size of 32, a learning rate of 2e-5, and a maximum context length of 64K.
For RL, starting from the SFT checkpoint, we train with GRPO on the 2,600 synthesized queries.
The training configuration includes a rollout size of 32, 8 samples per prompt, a global batch size of 256, a temperature of 1.0, a learning rate of 1e-6, and a maximum context length of 64K tokens.
Training is run for 2 epochs.
GPT-5.1~\cite{openai2025gpt51} serves as the judge LLM for both outcome reward scoring and evidence verification.
Full hyperparameters are listed in Appendix~\ref{sec:training_config}.

\paragraph{Baselines}
All controlled baselines share identical SFT initialization, training data, hyperparameters, and search engine; they differ only in reward design:
GRPO uses a binary outcome reward (0/1 correctness);
E-GRPO uses entity hit rate against the evidence graph as reward~\cite{zhao2026repurposingsyntheticdatafinegrained};
C-GRPO combines a rubric-based citation score with outcome correctness~\cite{zhang2026chainingevidencerobustreinforcement}.
Reward details are in Appendix~\ref{sec:training_config}.
For broader context, Table~\ref{tab:main_results} also lists proprietary and open-source agents as reference points.

\paragraph{Benchmarks}
We evaluate on BrowseComp~\cite{wei2025browsecompsimplechallengingbenchmark}, BrowseComp-ZH~\cite{zhou2025browsecompzhbenchmarkingwebbrowsing}, and xbench-DS~\cite{xbench2025xbench}.
Together, these benchmarks assess core capabilities for deep search: long-horizon information-seeking, multi-step web navigation, complex reasoning, and cross-lingual synthesis.
All models use the same inference environment (API, tool interface, 128K context window) with no turn or time limit.

\begin{table*}[t]
  \centering
  \small
  \renewcommand{\arraystretch}{1.15}
  \begin{tabularx}{\textwidth}{l >{\centering\arraybackslash}X >{\centering\arraybackslash}X >{\centering\arraybackslash}X >{\centering\arraybackslash}X}
    \toprule
    \textbf{Model} & \textbf{BrowseComp} & \textbf{\mbox{BrowseComp-ZH}} & \textbf{xbench-DS} & \textbf{Overall} \\
    \midrule
    \rowcolor{gray!10} \multicolumn{5}{c}{\textit{Advanced Agents with Proprietary Data}} \\
    \midrule
    Tongyi-DeepResearch~\cite{tongyideepresearchteam2026tongyideepresearchtechnicalreport} & 43.4 & 46.7 & 75.0 & 45.9 \\
    RedSearcher~\cite{chu2026redsearcherscalablecostefficientframework}         & 42.1 & 49.8 & - & - \\
    Gemini 3.1 Pro ~\cite{google}     & 53.6 & 75.9 & 86.0 & 59.5 \\
    Claude Sonnet 4.6 ~\cite{anthropic2026sonnet46}   & 29.5 & 42.4 & - & - \\
    Seed 2.0 Pro~\cite{seed2026seed20}   & 77.3 & 82.4 & 88.0 & 78.8 \\

    \midrule
    \rowcolor{gray!10} \multicolumn{5}{c}{\textit{Agents with Open-source Data}} \\
    \midrule
    Asearcher-Web-32B~\cite{gao2025turnsunlockinglonghorizonagentic}   & 5.2  & 15.6 & 42.1 & 9.2 \\
    WebSailor-32B ~\cite{Li2025WebSailorNS}      & 10.5 & 25.5 & 53.3 & 15.7 \\
    WebExplorer-8B~\cite{liu2025webexplorerexploreevolvetraining}        & 15.2 & 32.0 & 53.7 & 20.5 \\
    OpenSeeker-30B~\cite{du2026openseekerdemocratizingfrontiersearch}        & 29.5 & 48.4 & 74.0 & 35.5 \\
    \midrule
    \rowcolor{gray!10} \multicolumn{5}{c}{\textit{Controlled Comparison}$^\dagger$} \\
    \midrule
    Qwen-30B-SFT      & 11.1       & 20.4       & 57.0       & 15.5 \\
    \midrule
    + GRPO               & 18.9       & 30.8       & 70.0          & 24.1 \\
    \rowcolor{violet!8}
    + GRPO + \method{}   & \textbf{20.9} {\scriptsize(+2.0)}  & \textbf{36.3} {\scriptsize(+5.5)}  & \textbf{73.0} {\scriptsize(+3.0)}  & \textbf{26.7} {\scriptsize(+2.6)} \\
    \midrule
    + E-GRPO             & 15.9       & 27.7       & 65.0          & 20.9 \\
    \rowcolor{violet!8}
    + E-GRPO + \method{} & \textbf{17.0} {\scriptsize(+1.1)}  & \textbf{29.4} {\scriptsize(+1.7)}  & \textbf{65.0} {\scriptsize(+0.0)}  & \textbf{22.1} {\scriptsize(+1.2)} \\
    \midrule
    + C-GRPO             & 18.3       & 31.1       & 69.0          & 23.6 \\
    \rowcolor{violet!8}
    + C-GRPO + \method{} & \textbf{20.5} {\scriptsize(+2.2)}  & \textbf{34.6} {\scriptsize(+3.5)}  & \textbf{71.0} {\scriptsize(+2.0)}  & \textbf{26.0} {\scriptsize(+2.4)} \\
    \bottomrule
  \end{tabularx}
  \caption{Main results on three deep search benchmarks. $^\dagger$Controlled baselines share identical SFT initialization, training data, and search engine---differences are exclusively in reward design and credit attribution. }
  \label{tab:main_results}
  \vspace{-0.3cm}
\end{table*}

\subsection{Main Results}
\label{sec:main_results}

Table~\ref{tab:main_results} presents results across all three benchmarks.
All controlled methods start from the same SFT checkpoint; RL training with GRPO variants raises performance substantially across all reward designs. When \method{} is applied on top of the GRPO baselines, it yields consistent gains: +2.6 overall on GRPO, +1.2 on E-GRPO, and +2.4 on C-GRPO.
These gains come from injecting citation-traced step credit through sign-preserving advantage modulation---no additional reward signal is introduced---indicating that provenance-based credit assignment captures complementary optimization information beyond the trajectory-level reward.

The improvement holds across three base rewards with fundamentally different designs (outcome-only, entity-aware, and rubric-based), confirming that \method{}'s credit assignment is orthogonal to reward design and provides additive gains under the major reward formulations explored in recent deep-search agents.

\subsection{Ablation Studies}
\label{sec:ablation}

We conduct ablations on the outcome-only GRPO baseline to isolate the contribution of individual design choices in \method{}.

\paragraph{Evidence Verification Method}
Table~\ref{tab:ablation_verification} compares two approaches for determining whether a cited document supports target evidence.
String matching checks whether entity names and relation keywords appear literally in the document text; it captures surface-level co-occurrence but cannot handle paraphrases or verify semantic constraints.
Our method restricts verification to cited documents and uses an LLM verifier to judge entity and relation support, yielding substantially higher credit quality.

\begin{table}[h]
  \centering
  \small
  \renewcommand{\arraystretch}{1.15}
  \begin{tabularx}{\columnwidth}{>{\raggedright\arraybackslash}p{0.35\columnwidth} >{\centering\arraybackslash}X >{\centering\arraybackslash}X >{\centering\arraybackslash}X}
    \toprule
    \textbf{Verification Method} & \textbf{BC} & \textbf{BC-ZH} & \textbf{xbench} \\
    \midrule
    String matching       & 15.6 & 26.6 & 65.0 \\
    \textbf{URL + Judge (ours)} & \textbf{20.9} & \textbf{36.3} & \textbf{73.0} \\
    \bottomrule
  \end{tabularx}
  \caption{Ablation on evidence verification method.}
  \label{tab:ablation_verification}
  \vspace{-0.3cm}
\end{table}

\paragraph{Advantage Modulation Variants}
Table~\ref{tab:ablation_advantage} ablates three design dimensions.
``All-step'' distributes credit to every step that encounters a supporting URL rather than only the earliest, diluting the credit signal across redundant steps.
``Unbounded'' removes the per-step cap $C$, allowing a single evidence-rich step to dominate the gradient (sensitivity to $\delta$ and $C$ is examined in Appendix~\ref{sec:hyperparam_sensitivity}).
``Pos.\ only'' applies modulation exclusively to positive-advantage trajectories, removing the attenuation effect on negative ones.
Our full design (earliest attribution, bounded cap, both signs) achieves the best result.

\begin{table}[h]
  \centering
  \small
  \renewcommand{\arraystretch}{1.15}
  \resizebox{\columnwidth}{!}{%
  \begin{tabular}{lllccc}
    \toprule
    \textbf{Attribution} & \textbf{Modulation} & \textbf{Scope} & \textbf{BC} & \textbf{BC-ZH} & \textbf{xbench} \\
    \midrule
    All-step  & Bounded   & Both       & 19.2 & 31.1 & 71.0 \\
    Earliest  & Unbounded & Both       & 16.3 & 28.7 & 67.0 \\
    Earliest  & Bounded   & Pos.\ only & 20.1 & 34.0 & 71.0 \\
    \midrule
    \textbf{Earliest} & \textbf{Bounded} & \textbf{Both} & \textbf{20.9} & \textbf{36.3} & \textbf{73.0} \\
    \bottomrule
  \end{tabular}%
  }
  \caption{Design variant ablation. Each row changes one dimension from the full design (bottom).}
  \label{tab:ablation_advantage}
  \vspace{-0.3cm}
\end{table}

\subsection{Analysis}
\label{sec:analysis}

\begin{figure}[t]
  \centering
  \includegraphics[width=\columnwidth]{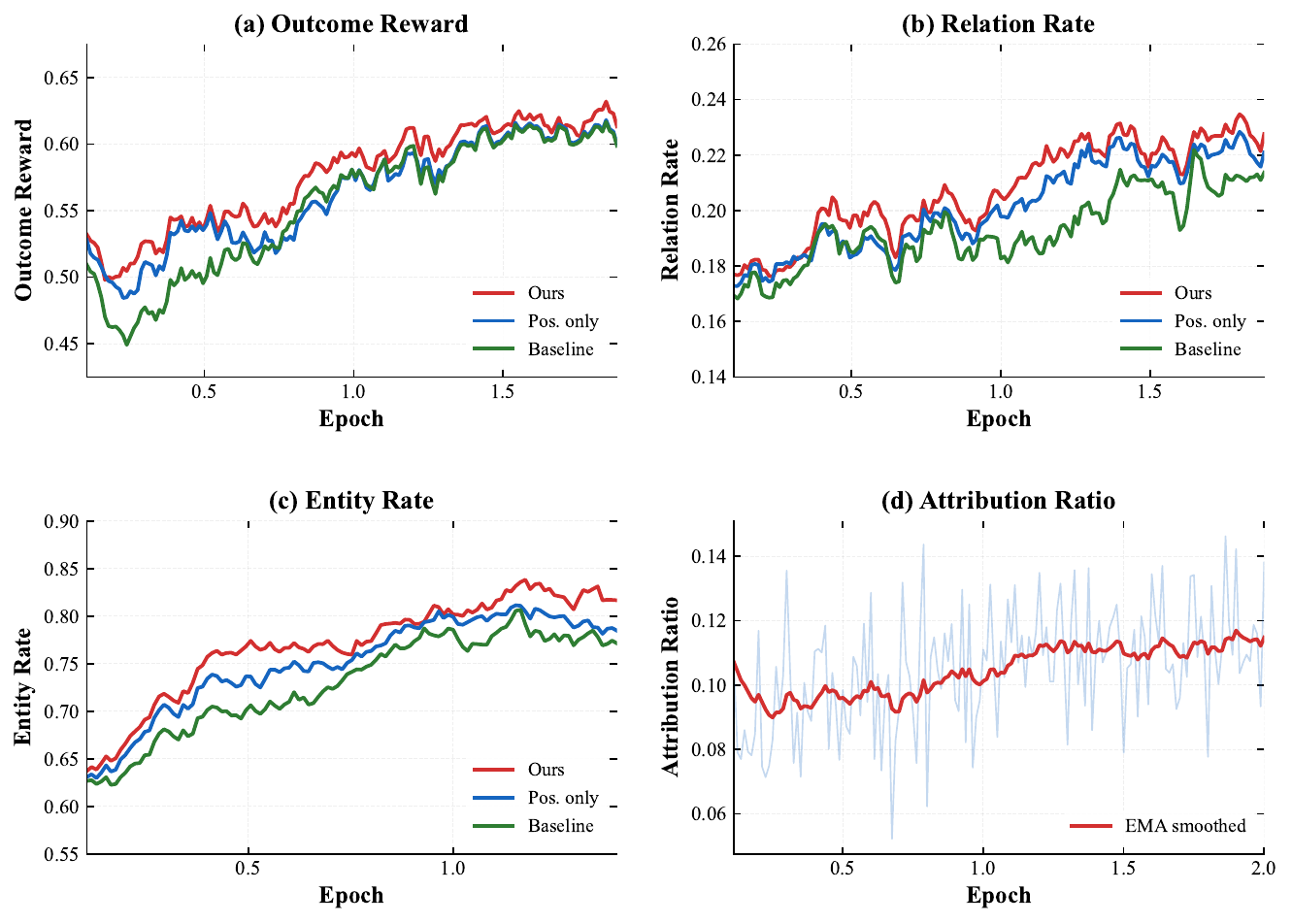}
  \caption{Training dynamics and credit sparsity. (a)~Outcome reward; (b)~relation verification rate; (c)~entity grounding rate; (d)~fraction of steps receiving non-zero credit (${\sim}10$--$12\%$), confirming that provenance-based attribution concentrates on a sparse subset of actions.}
  \label{fig:combined}
  \vspace{-0.5cm}
\end{figure}

\paragraph{Training Dynamics}
Figure~\ref{fig:combined} tracks outcome reward, relation verification rate, and entity grounding rate over training.
\method{} achieves higher outcome reward and converges faster than the GRPO baseline, indicating that localized credit accelerates policy improvement.
The separation is most pronounced on relation rate (Fig.~\ref{fig:combined}b): \method{} sustains a clear advantage throughout training, while the baseline plateaus at a lower level.
Entity rate (Fig.~\ref{fig:combined}c) saturates quickly with smaller separation---consistent with entity discovery being relatively easy, whereas relation verification requires targeted multi-hop evidence gathering that benefits most from step-level credit.

\paragraph{Credit Sparsity}
Figure~\ref{fig:combined}d shows the fraction of steps receiving non-zero credit over training.
Throughout the process, only ${\sim}10$--$12\%$ of steps are credited, confirming that provenance-based attribution concentrates optimization pressure on a small subset of actions rather than spreading it uniformly.
The slight upward trend (${\sim}9\%$ at epoch~0.3 to ${\sim}11.5\%$ at epoch~2.0) suggests a positive feedback loop: as the policy improves, it issues more evidence-producing queries, which generate more credit signal that further reinforces targeted evidence gathering.

\section{Related Work}

\paragraph{RL and Reward Design for Deep Search Agents}
Reinforcement learning has become central to training deep search agents that interact with live web environments~\cite{jin2025searchr1trainingllmsreason}.
Recent work improves both the data and infrastructure for agentic RL~\cite{lu2025deepdiveadvancingdeepsearch,tao2025webshaperagenticallydatasynthesizing} and the reward signals used to supervise multi-turn search behavior~\cite{zhang2025processvsoutcomereward}.
Beyond binary outcome rewards, citation-aware rubrics, entity-level rewards, and decomposed evidence constraints provide denser trajectory-level supervision.
CaRR~\cite{zhang2026chainingevidencerobustreinforcement} is closest to our setting: it decomposes questions into single-hop constraints and verifies cited evidence with an LLM-based rubric.
E-GRPO~\cite{zhao2026repurposingsyntheticdatafinegrained} similarly rewards entity match rate to distinguish near-miss rollouts from complete failures.
These methods refine \emph{what} a trajectory should be rewarded for, but the resulting scalar reward is still broadcast uniformly across actions; they do not identify which search or read step first exposed the verified evidence.

\paragraph{Step-Level Credit Assignment}
Temporal credit assignment is a foundational challenge in reinforcement learning~\cite{pignatelli2024surveytemporalcreditassignment}.
Process reward models provide local supervision in mathematical reasoning~\cite{zhang2025lessonsdevelopingprocessreward}, but open-web search lacks canonical labels for arbitrary intermediate actions and evolves over time.
Hindsight relabeling and token-level reward modeling offer alternative ways to densify sparse feedback~\cite{andrychowicz2018hindsightexperiencereplay}, yet they either operate over goal relabeling or require learned local reward models.
\method{} instead derives step-level credit from citation provenance already present in the trajectory: a reference-based verifier determines which cited documents support evidence units, and first-exposure attribution maps those supported documents back to the actions that surfaced them.
This provides localized optimization pressure without changing the trajectory-level reward or requiring additional human step annotations.

\section{Conclusion}
\label{sec:conclusion}

We presented \textsc{STAMP}, a citation-provenance approach for assigning step-level credit in deep-search RL.
The central idea is to separate trajectory-level scoring from step-level credit assignment: a reference-based verifier identifies which cited documents support target entities and constraints, citation provenance maps those supporting documents to their first-exposure actions, and sign-preserving advantage modulation injects the resulting credit without changing the trajectory-level reward or its induced ranking.

Across BrowseComp, BrowseComp-ZH, and xbench-DS, \textsc{STAMP} consistently improves controlled GRPO baselines under matched training and search conditions, with gains of \textbf{+2.0}/\textbf{+5.5}/\textbf{+3.0} points over the controlled baseline on each benchmark.
Ablations further show that verification quality, first-exposure attribution, bounded modulation, and negative-advantage attenuation are all important to the final performance.
These results indicate that citation provenance carries optimization signal not captured by trajectory-level rewards alone.

\section{Limitations}
All experiments are conducted on Qwen3-30B-A3B-Thinking.
While this scale is representative of current open-source deep-search agents, we have not validated \method{} on substantially larger backbones (e.g., 70B+ dense or MoE models).
Whether provenance-guided credit remains equally beneficial as model capacity grows is an open question.

The open-source community has released various search-augmented SFT checkpoints with differing training recipes and tool interfaces.
To ensure fully controlled comparison---identical SFT data, tool definitions, search environment, and inference pipeline---we initialize exclusively from our own SFT checkpoint rather than adopting external ones.
Extending \method{} to diverse community checkpoints is a natural next step.

\bibliography{custom}

\appendix

\section{Data Synthesis Details}
\label{sec:chp_details}

\subsection{Tree Structure Examples}

\paragraph{Easy ($n_\text{hidden}=1$, 2 leaves, 1 inference step)}
Two leaf nodes converge to uniquely identify the answer:
\begin{verbatim}
Leaf_A (attended) --> D (answer)
Leaf_B (graduated) --> D (answer)
\end{verbatim}
Agent must find what both Leaf\_A and Leaf\_B share.

\paragraph{Medium ($n_\text{hidden}=2$, 3--5 leaves, 2 inference steps)}
\begin{verbatim}
Leaf_A (connects) --> Hidden_P
Leaf_B (hosted)   --> Hidden_P
Hidden_P (located_in) --> D
Leaf_C (adjacent_to)  --> D
\end{verbatim}
Step~1: Identify Hidden\_P from Leaf\_A and Leaf\_B.
Step~2: Combine Hidden\_P with Leaf\_C to determine D.

\paragraph{Hard ($n_\text{hidden}\geq4$, 5+ leaves, 3+ inference steps)}
Multiple layers of hidden nodes requiring recursive inference across 3+ steps.

\subsection{Relation Refinement}
\label{sec:relation_refine}
Raw relation constraints are refined to ensure each constraint references exactly two entity placeholders.
Constraints spanning 3+ entities are split into atomic binary relations;
constraints referencing only one entity are merged with related constraints.
This normalization ensures consistent granularity across questions of different complexity.

\subsection{Edge Verification Details}
The dual-path judge evaluates three criteria for each edge:
(1)~whether the stated relation factually holds;
(2)~whether the relation is current (not outdated);
(3)~whether evidence from both paths is consistent.
This catches common failure modes in raw KG edges: outdated sports affiliations, retracted mergers reported as completed, and editorial ``See Also'' links misinterpreted as semantic relations.

\section{Training Configuration}
\label{sec:training_config}

\subsection{Model Architecture}
We use Qwen3-30B-A3B-Thinking-2507 as the base model: 30B total parameters with 3B active per token via Mixture-of-Experts routing.
Supports up to 128K context; we use 64K maximum context length during both SFT and RL training.
Both stages are implemented on the VeRL~\cite{sheng2024hybridflow} training framework.

\subsection{SFT Stage}
We use Seed~Pro~2.0 to generate 1,500 high-quality search trajectories on the SFT query set, then fine-tune the base model for 1 epoch with batch size 32, learning rate 2e-5, and maximum context length 64K.
Training is conducted on 16$\times$H800 GPUs.
All Training baselines share this SFT initialization.

\subsection{RL Stage}
Starting from the SFT checkpoint, we train with GRPO on 2,600 synthesized queries using 16$\times$H800 GPUs.
GPT-5.1 serves as the verification judge for evidence reward computation.
Table~\ref{tab:rl_hyperparams} lists the complete hyperparameters.

\begin{table}[htbp]
  \centering
  \small
  \setlength{\tabcolsep}{6pt}
  \renewcommand{\arraystretch}{1.2}
  \begin{tabular}{ll}
    \toprule
    \textbf{Hyperparameter} & \textbf{Value} \\
    \midrule
    Training questions & 2.6K (2.1K EN + 0.5K ZH) \\
    Rollouts per prompt & 8 \\
    Rollout batch size & 32 \\
    Global batch size & 256 \\
    Sampling temperature & 1.0 \\
    Maximum context length & 64K tokens \\
    Training epochs & 2 \\
    Learning rate & $1 \times 10^{-6}$ \\
    KL coefficient ($\beta$) & 0.01 \\
    Clip range ($\varepsilon$) & 0.2 \\
    Credit quantum $\delta$ & 0.03 \\
    Per-step credit cap $C$ & 0.10 \\
    Max cited URLs for verification & 15 \\
    Verification model & GPT-5.1 \\
    \bottomrule
  \end{tabular}
  \caption{Complete RL training hyperparameters.}
  \label{tab:rl_hyperparams}
\end{table}

\subsection{Training Data Composition}
The 2.6K RL training questions are distributed across difficulty levels:
easy ($n_\text{hidden}{=}1$): 40\%,
medium ($n_\text{hidden}{=}2$--$3$): 45\%,
hard ($n_\text{hidden}{\geq}4$): 15\%.
This distribution reflects the natural yield from \datasyn{} mining after verification filtering.

\section{\method{} Training Algorithm}
\label{sec:algorithm}

Algorithm~\ref{alg:prism} summarizes the complete \method{} training loop. Each iteration rolls out a group of trajectories, verifies cited URLs against the evidence graph, attributes credit to first-exposure steps, and applies sign-preserving modulation before the policy update.

\begin{algorithm}[htbp]
\caption{\method{} training: reward-decoupled citation-traced credit with sign-preserving modulation}
\label{alg:prism}
\begin{algorithmic}[1]
\Require Policy $\pi_\theta$, training queries $\{(q, \mathcal{G}, \Vtar, \Utar)\}$, base reward $R(\cdot)$, group size $G$, credit quantum $\delta$, cap $C$
\For{each training iteration}
  \State Sample query $(q, \mathcal{G}, \Vtar, \Utar)$ with training-time evidence graph $\mathcal{G} = (\mathcal{V}, \mathcal{E})$
  \State Roll out $G$ trajectories $\{\tau_i\}_{i=1}^G$ from $\pi_\theta$ given $q$
  \For{each $\tau_i$}
    \State \textbf{Verify:} run verification (\S\ref{sec:reward}) against $\mathcal{G}$ using cited URLs in $\tau_i$, producing $\{\mathcal{D}(e)\}, \{\mathcal{D}(u_k)\}, \{\supp(e)\}, \{\verifiedop(u_k)\}$
    \State Compute base reward $R_i = R(\tau_i)$
    \State \textbf{Attribute:} distribute $\delta$ to supporting documents via entity/relation channels, trace to steps via $\stepof(d)$ (Eq.~\ref{eq:step_attr})
    \State Accumulate $\credit_s^{(i)}$ per action step with cap $C$ (Eq.~\ref{eq:step_credit})
  \EndFor
  \State Compute group advantages $A_i = (R_i - \mu)/(\sigma + \epsilon)$ (Eq.~\ref{eq:grpo_advantage})
  \State Compute step-modulated $\widetilde{A}_{i,s}$ via $\mathcal{M}_{\credit_s^{(i)}}(A_i)$ (Eq.~\ref{eq:modulation_op}) and broadcast to action tokens as $A^*_{i,m}$
  \State Update $\theta$ by minimizing $\mathcal{L}(\theta)$ (Eq.~\ref{eq:stamp_loss})
\EndFor
\end{algorithmic}
\end{algorithm}

\section{Supplementary Experiments}
\label{sec:hyperparam_sensitivity}

Table~\ref{tab:sensitivity} examines sensitivity to the credit quantum $\delta$ and per-step cap $C$.
Performance peaks at $\delta{=}0.03$ and $C{=}0.10$; smaller values underweight the credit signal, while larger values concentrate too much advantage on a few steps.

\begin{table}[h]
  \centering
  \small
  \renewcommand{\arraystretch}{1.15}
  \begin{tabularx}{\columnwidth}{>{\centering\arraybackslash}X >{\centering\arraybackslash}X >{\centering\arraybackslash}X >{\centering\arraybackslash}X}
    \toprule
    $\delta$ & \textbf{BC} & \textbf{BC-ZH} & \textbf{xbench} \\
    \midrule
    0.01 & 17.8 & 29.4 & 69.0 \\
    \textbf{0.03 (ours)} & \textbf{20.9} & \textbf{36.3} & \textbf{73.0} \\
    0.05 & 19.7 & 33.2 & 70.0 \\
    \midrule
    $C$ & \textbf{BC} & \textbf{BC-ZH} & \textbf{xbench} \\
    \midrule
    0.05 & 17.9 & 29.1 & 71.0 \\
    \textbf{0.10 (ours)} & \textbf{20.9} & \textbf{36.3} & \textbf{73.0} \\
    0.15 & 16.3 & 27.7 & 65.0 \\
    \bottomrule
  \end{tabularx}
  \caption{Sensitivity to credit quantum $\delta$ and per-step cap $C$.}
  \label{tab:sensitivity}
\end{table}

\section{Human Verification of LLM Judge}
\label{sec:human_verification}

To assess the reliability of LLM as the verification judge for evidence reward computation, we conducted a manual review of its judgments across 200 randomly sampled training trajectories from the SFT checkpoint.
Using human assessments as the gold standard, the judge LLM achieved an accuracy of 96.04\% for entity identification (whether a cited document supports the presence of a target entity) and 97.31\% for relation verification (whether a cited document confirms the stated relationship between entities).
The majority of disagreements involved borderline cases where evidence was implicit rather than explicitly stated.
These results indicate that the LLM judge provides sufficiently reliable verification signals for credit attribution during training.

\section{Prompts and Formats}
\label{sec:prompts}

\subsection{Tool Descriptions}
\label{sec:tool_desc}

The agent is equipped with two tools for interacting with the live web. Their specifications are provided verbatim in the system prompt:

\begin{figure*}[t]
\begin{promptbox}{Tool Definitions}
\begin{verbatim}
[
  {
    "name": "search",
    "description": "Search the web for information. Returns top results with url and snippet.",
    "parameters": {
      "type": "object",
      "properties": {
        "query": {"type": "string", "description": "The search query."},
        "num": {"type": "integer", "description": "Number of results (default 10)."}
      },
      "required": ["query"]
    }
  },
  {
    "name": "open",
    "description": "Open a page by URL and return the complete content.",
    "parameters": {
      "type": "object",
      "properties": {
        "id": {"type": "string", "description": "The absolute URL from prior search results."}
      },
      "required": ["id"]
    }
  }
]
\end{verbatim}
\end{promptbox}
\caption{Tool definitions provided to the agent in the system prompt.}
\label{fig:tool_def}
\end{figure*}


\subsection{Trajectory Format}
\label{sec:traj_format}

Each training trajectory follows the multi-turn format below. The agent alternates between reasoning (within \texttt{<think>} tags), tool invocations, and environment observations until producing a final cited response.

\begin{figure*}[t]
\begin{promptbox}{Trajectory Format}
\begin{verbatim}
<|im_start|>system
You are a helpful assistant. ...
<tools> [tool definitions] </tools>
<|im_end|>
<|im_start|>user
[question + instruction]
<|im_end|>
<|im_start|>assistant
<think> reasoning process </think>
<tool_call>
{"name": "search", "arguments": {"query": "..."}}
</tool_call>
<|im_end|>
<|im_start|>user
<tool_response> search results </tool_response>
<|im_end|>
<|im_start|>assistant
<think> reasoning process </think>
<tool_call>
{"name": "open", "arguments": {"id": "https://..."}}
</tool_call>
<|im_end|>
<|im_start|>user
<tool_response> page content </tool_response>
<|im_end|>
... (more search/open iterations) ...
<|im_start|>assistant
<think> final reasoning </think>
## Complete Answer
[explanation with inline citations, e.g. [1], [2]]
## Exact Answer
[final answer]
## References
[1] https://...
[2] https://...
<|im_end|>
\end{verbatim}
\end{promptbox}
\caption{Multi-turn trajectory format used during training and inference.}
\label{fig:traj_format}
\end{figure*}


\subsection{Prompt for Outcome Rewards}
\label{sec:prompt_outcome}

The outcome reward is computed by prompting GPT-5.1 to judge whether the agent's final answer matches the ground-truth answer. The judge extracts the answer from the response, compares it against the reference, and returns a binary correctness label (Figure~\ref{fig:prompt_outcome}).

\begin{figure*}[t]
\begin{promptbox}{Prompt for Outcome Rewards}
Judge whether the following \textbf{[response]} to \textbf{[question]} is correct or not based on the precise and unambiguous \textbf{[correct\_answer]} below.

\medskip
\textbf{[question]:} \{question\}

\textbf{[response]:} \{response\}

\medskip
Your judgement must be in the format and criteria specified below:

\medskip
\textbf{extracted\_final\_answer:}
The final exact answer extracted from the \textbf{[response]}. Put the extracted answer as `None' if there is no exact, final answer to extract from the response.

\medskip
\textbf{[correct\_answer]:} \{correct\_answer\}

\medskip
\textbf{reasoning:}
Explain why the \textbf{extracted\_final\_answer} is correct or incorrect based on \textbf{[correct\_answer]}, focusing only on whether there are meaningful differences between \textbf{[correct\_answer]} and the \textbf{extracted\_final\_answer}. Do not comment on any background to the problem. Do not attempt to solve the problem. Focus only on whether the answers match.

\medskip
\textbf{correct:}
Answer `yes' if the \textbf{extracted\_final\_answer} matches the \textbf{[correct\_answer]} given above, or is within a small margin of error for numerical problems. Answer `no' otherwise, including cases of inconsistency, ambiguity, non-equivalency, or incorrectness.
\end{promptbox}
\caption{Prompt for outcome reward judgment.}
\label{fig:prompt_outcome}
\end{figure*}


\subsection{Prompt for Entity Verification}
\label{sec:prompt_entity}

Entity verification (Stage~1 of evidence reward) checks whether each target entity is correctly identified in the response and supported by cited webpage evidence. The verifier receives the question, relation constraints with entity placeholders, ground-truth values, the agent's response, and the cited webpage contents (Figure~\ref{fig:prompt_entity}).

\begin{figure*}[t]
\begin{promptbox}{Prompt for Entity Verification}
You will receive:
\begin{enumerate}[leftmargin=*, nosep]
\item A complex multi-hop question.
\item A set of relation constraints with entity placeholders (\texttt{<E0>}, \texttt{<E1>}, \ldots).
\item The correct (ground truth) values for each entity.
\item An AI assistant's response to the question.
\item The contents of webpages cited by the assistant (evidence).
\end{enumerate}

\medskip
\textbf{Your task} (for each entity):
\begin{enumerate}[leftmargin=*, nosep]
\item Extract the identity of each entity from the assistant's response. For \texttt{<E0>}, prioritize the \texttt{\#\# Exact Answer} section. For other entities, look in the \texttt{\#\# Explanation with Citations} section.
\item Judge whether each extracted entity value matches its ground truth value. Consider equivalent names, minor wording differences, abbreviations, or alternate transliterations as correct.
\item Judge whether each entity's identity is explicitly mentioned or supported by the provided webpage evidence. If supported, indicate which URL contains the evidence.
\end{enumerate}

\medskip
\textbf{[Question]:} \{question\}

\textbf{[Relation Constraints]:} \{constraints\}

\textbf{[Ground Truth Entity Values]:} \{ground\_truth\}

\textbf{[Assistant's Response]:} \{response\}

\textbf{[Webpage Contents (cited evidence)]:} \{context\}

\medskip
\textbf{Output format:}
\begin{verbatim}
{
  "E0": {"identified_value": "...", "correct": true/false,
         "supported": true/false, "source_url": "URL N"},
  "E1": {...},
  ...
}
\end{verbatim}
\end{promptbox}
\caption{Prompt for entity verification.}
\label{fig:prompt_entity}
\end{figure*}


\subsection{Prompt for Relation Verification}
\label{sec:prompt_relation}

Relation verification (Stage~2 of evidence reward) determines whether each relation constraint between identified entities is supported by the cited webpage contents. Unlike entity verification which checks identity, this stage evaluates whether the stated factual relationship can be confirmed from the evidence (Figure~\ref{fig:prompt_relation}).

\begin{figure*}[t]
\begin{promptbox}{Prompt for Relation Verification}
You will receive:
\begin{enumerate}[leftmargin=*, nosep]
\item The contents of several webpages (evidence collected by the AI assistant).
\item Several relation constraints: S1, S2, \ldots, Sn --- factual statements describing relationships between identified entities.
\end{enumerate}

\medskip
\textbf{Your task:} For each constraint:
\begin{itemize}[leftmargin=*, nosep]
\item Find the exact evidence from the webpage contents that supports or contradicts it.
\item A constraint is ``supported'' only if the webpage content contains sufficient information to verify the stated relationship or fact.
\item Conclude with a judgment: \textbf{Fully Supported: yes} or \textbf{Fully Supported: no}.
\end{itemize}

\medskip
\textbf{[Webpage Contents]:} \{context\}

\textbf{[Relation Constraints]:} \{statements\}

\medskip
\textbf{Output format:}
\begin{verbatim}
{
  "S1": {"supported": true/false, "source_url": "URL N"},
  "S2": {"supported": true/false, "source_url": "URL N"},
  ...
}
\end{verbatim}
\end{promptbox}
\caption{Prompt for relation verification.}
\label{fig:prompt_relation}
\end{figure*}
\end{document}